# Shortcut Learning in a Public Grape Disease Dataset: Annotation Granularity as a Modulator, Not a Cause

**Author** Pushuo Wang

**Affiliation** School of Information and Control Engineering, Shenyang Institute of Technology, Fushun, Liaoning, China

## Abstract

Public datasets are the main data source for research on agricultural disease detection, and their fitness for use is usually judged from the metrics reported on them — yet those metrics say nothing about whether the annotation scheme is internally consistent. Taking one public grape disease detection dataset (3288 images, 11995 boxes, 6 classes) as the object of study, this paper asks which intrinsic properties of public data determine whether a detection system built on it is actually usable. Five single-variable controls spanning model capacity from 2.58 M to 32 M parameters, input resolution from 384 to 960, and one change of detection paradigm yield a test-set mAP50 range of only 1.54 percentage points under a single evaluation protocol — the same order of magnitude as seed-to-seed variation; scale-stratified evaluation locates the bottleneck at small objects consistently across all five architectures. The principal finding lies on the data side: one class, *mosaic virus disease*, is annotated predominantly at whole-leaf level (median box area 43.16% of the image), while the other five are annotated consistently at lesion level (0.57% to 8.12%). Using 5156 cross-species images containing no grape as a negative control, 65.7% of the 12221 false-positive boxes the model emits fall into that one class — an over-representation of 13.41× relative to its share of the training annotations, in the direction opposite to a class-frequency prior. Single-variable counterfactual retraining establishes a causal effect of granularity on the **magnitude** of the shortcut: shrinking only that class's boxes lowers its in-distribution AP from 0.7609 to 0.4552 while the other five classes change by no more than seed-to-seed variation, and cross-species false positives fall by 66% at the same time; a placebo control confirms that the effect is specific to the manipulated class. A manipulation in the opposite direction, with criteria registered in advance, returns a negative result: coarsening the most finely annotated class to whole-leaf level (median box area 0.57% → 40.37%), so that it matches the first class in granularity, box count and share of annotations, leaves its cross-species false positives at zero boxes (highest confidence 0.009, in-distribution mAP50 0.990), while the unmanipulated original class still accounts for 50.0% of all false positives. The claim is therefore narrowed to this: **annotation granularity determines the magnitude of the shortcut, not its destination** — granularity can amplify or attenuate a sink that already exists, but is not sufficient to create one; what determines the destination remains undetermined. We further propose a granularity screening statistic that requires neither images nor training, and use it under a single protocol to separate two regimes — *uniformly coarse* and *uneven across classes* — of which only the latter is dangerous; note that this statistic measures whether the annotation scheme is internally consistent, not where false positives will go. Separately, an imaging-optics calculation shows that airborne detection of millimetre-scale lesions is unattainable at any operationally feasible altitude. The failure mode reported here is invisible to in-distribution evaluation, and therefore constitutes a blind spot in judging whether a public dataset is fit for use.

# 1 Introduction

## 1.1 Background

The major fungal diseases of grape have a latent period: infection completes within a particular window of temperature and humidity, while visible symptoms appear one to several weeks later. The effective window for control therefore does not coincide with the moment at which symptoms are observed.

The author has kept a household grape plot under continuous observation. In previous years the plot was treated with a single application of fungicide at the young-fruit stage, with timing and frequency decided from experience and unsupported by monitoring data. Although the spraying was carried out as planned, disease still appeared roughly one month before ripening, presenting as blackening of the fruit surface.

The plot was left untreated this year. In late July, pinpoint black spots appeared on the fruit surface (the berries were then green and had not begun to colour); by early August most berries on the same cluster carried such spots, some showed round dark lesions with a sunken centre, and others showed half-berry or whole-berry browning; brown necrotic spots and marginal leaf scorch were visible on the foliage. The causal pathogen has not been confirmed in the laboratory, and this paper therefore makes no determination of it.

What this sequence demonstrates is that **the decision of when to spray depends on observing symptoms, and by the time symptoms appear the infection is already complete.** Early, objective, quantifiable observation of disease signs is therefore a precondition for improving the timing of control, and was the original motivation for this project.

Pergola and trellis training spread the canopy above head height. The oily yellow blotches of downy mildew and the white mycelial layer of powdery mildew appear mainly on the upper leaf surface, whereas a person standing under the canopy sees only the underside. This is precisely where low-altitude nadir observation has value relative to manual ground inspection.

It should be stated that all experiments in this paper use ground-level close-range images; unmanned aerial vehicles are not an experimental object here. Flight-related conclusions appear only in Section 5.1 and rest solely on imaging optics.

## 1.2 Problem

This paper is organised around one overarching question: **which intrinsic properties of a public dataset determine whether a detection system trained on it is actually usable?** The question decomposes into three sub-questions, each answered by one part of the paper: what existing methods can achieve on this data (Section 3); if they fall short, whether the constraint lies in model capacity, in input information, or in the data itself (Sections 3.2 and 3.3); and whether high in-distribution metrics can be taken as evidence of practical usability (Section 4). The acquisition feasibility discussed in Section 5.1 is an extension of the same question to the acquisition stage — before data can be annotated, it must first exist optically.

The third sub-question is the focus of this paper. It matters because getting it wrong does not announce itself: the model keeps producing output, the format is normal, the confidences are plausible, and the conclusions are wrong.

## 1.3 Contributions

- Five single-variable controls under a single evaluation protocol, establishing the performance ceiling attainable on this data, together with a five-seed repetition that supplies the noise baseline any between-group comparison requires;
- Scale-stratified evaluation reproduced across five architectures, locating the bottleneck at small objects and showing that the improvement there is masked by aggregate metrics;
- Quantification of the inconsistency of annotation granularity across classes, and a cross-species negative control on 5156 images that verifies the shortcut learning it induces;
- Four single-variable counterfactual retrainings that separate what annotation granularity does: a shrink group and a placebo control establish its causal effect on the **magnitude** of the shortcut, while a manipulation in the opposite direction (coarsening the finest class to whole-leaf level), with criteria registered in advance, returns a negative result — narrowing the claim to "granularity determines magnitude, not destination";
- A granularity screening statistic $R$ computable without images or training, used under a single protocol to separate two regimes — *uniformly coarse* and *uneven across classes* — on two datasets;
- A calculation of the optical feasibility boundary for airborne lesion-level detection.

## 1.4 Related work

**Shortcut learning.** The phenomenon in which a model scores highly within the training distribution while relying on superficial cues unrelated to the task has been systematised as shortcut learning[1]. Its characteristic signature is normal in-distribution performance together with sharp out-of-distribution degradation, undetectable from aggregate metrics. The classic case in medical imaging is a model that decides on the basis of in-hospital markers on a chest radiograph rather than on the lesion[2]; in natural images there are reports of models that recognise objects from scene context rather than from the object itself[3]. Work on attribution methods likewise shows that some models with excellent benchmark performance in fact rely on irrelevant regions of the image[4].

This paper differs from prior work in three respects. First, most reported cases concern image classification; shortcuts in object detection are less often reported, and in detection the *scale of the annotation box* is itself an exploitable signal — a channel that classification does not have. Second, most studies attribute shortcuts to **image content** — background, watermarks, traces of the capture device — whereas the cause located here lies in the **annotation scheme**: the images themselves are unremarkable, every individual box is correct, and the problem is the inconsistency of annotation granularity *between* classes. Third, most work stops at describing the phenomenon and attributing it post hoc; here, single-variable counterfactual retraining (Section 4.4) establishes a causal effect of annotation granularity on the magnitude of the shortcut, a placebo control verifies the specificity of the effect, and a reverse manipulation marks the boundary of that attribution — granularity does not determine where the shortcut goes.

**Dataset annotation quality.** Mainstream benchmark datasets contain appreciable numbers of label errors, and those errors are sufficient to change model rankings[5]; re-examination of benchmark annotations further shows that the original annotation protocol itself can limit the validity of evaluation[6]. What concerns this paper is not label **error** — every box in the dataset used here encloses a genuine disease sign — but label **inconsistency**: within one dataset, one class is annotated predominantly at whole-leaf level while the other five are annotated at lesion level (Section 2.1). Such a problem cannot be caught by label-correction methods, because there is no incorrect label, only non-uniform granularity; it is equally invisible in in-distribution evaluation, where it in fact manifests as that class having the *highest* AP (Table 7). To our knowledge, shortcuts induced by inconsistent annotation granularity within a detection dataset have not previously been reported systematically.

**Visual recognition of plant disease.** Public data in this field are dominated by PlantVillage and datasets derived from it[7], on which reported classification accuracies are generally very high[8]. Subsequent work has pointed out, however, that such data are largely captured against controlled backgrounds, that a model can complete the classification from background information alone, and that the high accuracy therefore does not reflect capability under field conditions[9]. Datasets captured under field conditions have since been released to mitigate this[10].

Existing criticism is directed mainly at **acquisition conditions** (controlled backgrounds, detached single leaves, uniform illumination); this paper is directed at the **annotation scheme**. The two are independent dimensions of defect. The grape disease dataset used here is itself mixed in acquisition conditions — it contains studio-style close-ups with blurred backgrounds, field photographs of hand-held diseased clusters, and images composited from several photographs (Section 2.1) — yet the inconsistency of annotation granularity runs through all of them. It follows that improving acquisition conditions does not remove the problem described here: if a non-uniform annotation standard is carried over when a dataset is rebuilt, the same shortcut is carried into the new data.

**Small object detection and evaluation protocol.** Detection accuracy on small objects is markedly lower than on large ones, and this gap is stable across architectures[11]; sliced inference is one common mitigation[12]. On the evaluation side, existing work has argued that aggregate metrics mask specific failure modes and has advocated decomposing detection error by type[13]. Section 3.3 below gives an instance of the same problem along the **scale dimension**: raising the input resolution improves small-object AP significantly ($p = 0.018$) while aggregate mAP50-95 does not change ($p = 0.731$), so that relying on the latter alone leads to a conclusion contrary to fact. Section 5.1 further shows that in airborne settings this problem cannot be mitigated by any detection method — when a target occupies less than one pixel on the sensor, its information is already lost at acquisition.

# 2 Data and experimental setup

## 2.1 Dataset

The main experiments use the grape disease detection dataset **grape** published on Roboflow Universe (workspace `wscs`, project identifier `grape-uyimv`, version 1, released 2025-11-03, licensed CC BY 4.0, at https://universe.roboflow.com/wscs/grape-uyimv ), comprising 3288 images, 11995 annotation boxes and 6 classes, split 2631/329/328. Images were preprocessed by the platform to 640×640 (stretch resize after EXIF orientation stripping) with no image augmentation applied. The annotation target is lesion-level objects, averaging 3.65 boxes per image. The dataset is published by a platform user without a named author; in accordance with CC BY 4.0, the address above is given as the attribution.

**Table 1 Class distribution**

| Class | Boxes | Share | Principal symptom morphology |
|---|---|---|---|
| grape black rot | 4237 | 35.3% | Brown round spots, sharply bounded |
| grape powdery mildew | 2329 | 19.4% | White powdery mycelium in patches |
| grape downy mildew | 1740 | 14.5% | Oily yellow blotches on the leaf |
| grape botrytis cinerea | 1553 | 12.9% | Grey mould infecting clusters and stems |
| grape ulcer disease | 1543 | 12.9% | Browning and necrosis of stem tissue |
| mosaic virus disease | 593 | 4.9% | Whole-leaf mottling and distortion |

The ratio of the largest to the smallest class is 7.1:1, within an acceptable range, so all classes take part in training without resampling.

Computing the area of each annotation box as a fraction of the whole image reveals one conspicuous difference:

**Table 2 Box area as a fraction of the image, by class**

| Class | Median | Mean | P90 |
|---|---|---|---|
| mosaic virus disease | **43.16%** | 36.08% | 64.85% |
| grape botrytis cinerea | 8.12% | 13.02% | 29.47% |
| grape ulcer disease | 3.29% | 3.96% | 6.19% |
| grape downy mildew | 2.39% | 4.74% | 8.87% |
| grape powdery mildew | 2.01% | 2.95% | 4.93% |
| grape black rot | 0.57% | 0.69% | 1.21% |

The median box area of mosaic virus disease is 5.3 times that of the second-largest class and 76 times that of the smallest.

Examining the shape of that class's distribution further shows that its annotation granularity is **not uniform within the class either**: of its 593 boxes, 58% exceed 30% of the image (corresponding to whole leaves) and 28% fall below 10% (corresponding to local lesions), with quantiles P10 = 5.3%, P25 = 8.9%, median 43.2%, P75 = 57.6% — a bimodal shape. By comparison, only 10% of grape botrytis cinerea boxes exceed 30%.

The accurate statement is therefore this: that class is annotated predominantly at whole-leaf level with lesion-level boxes mixed in, while the other five are annotated consistently at lesion level. Inspection of the images confirms this. All of the analysis in Section 4 proceeds from here.

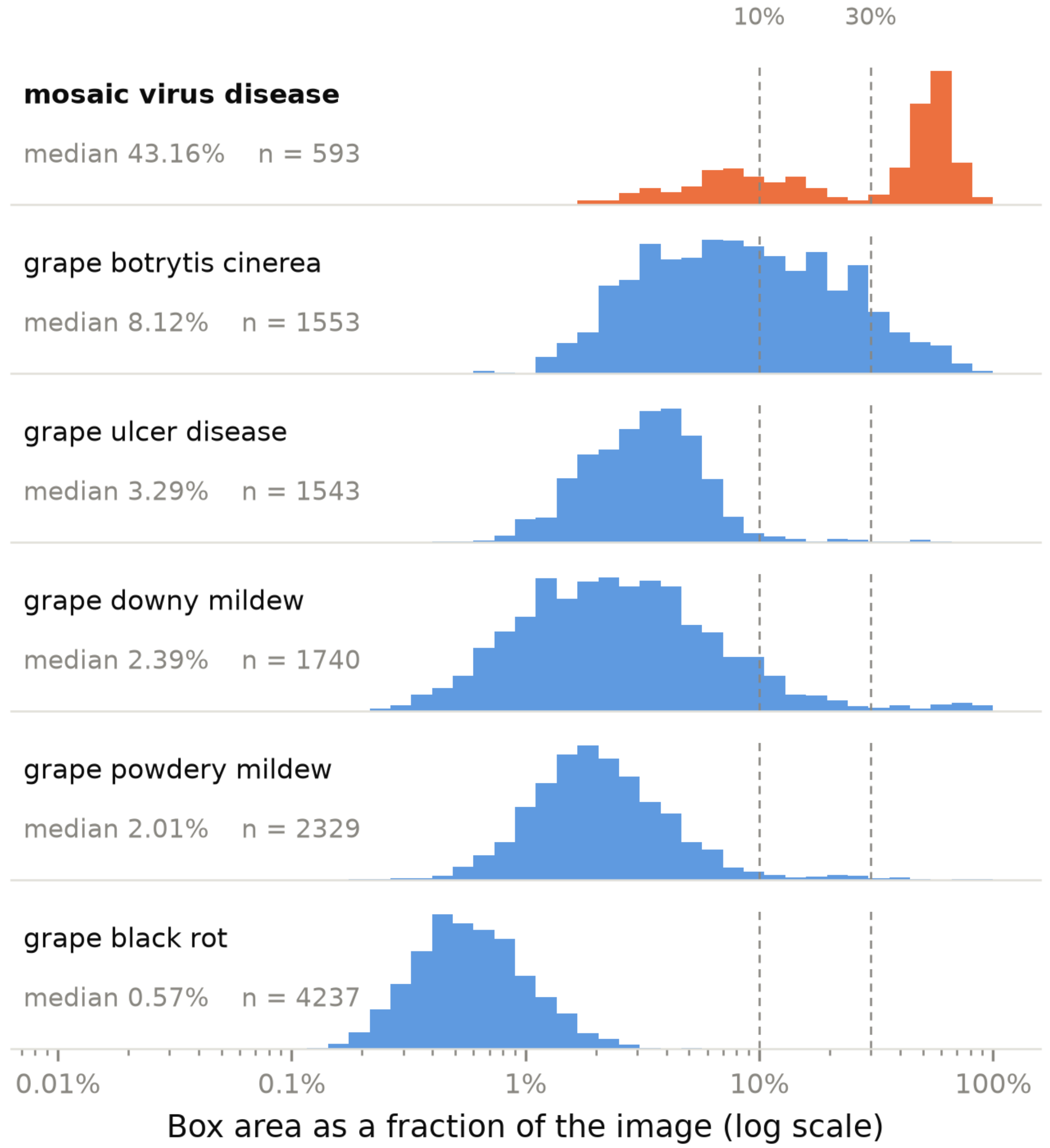

**Figure 1** Distribution of relative box area for the six classes (logarithmic abscissa; dashed lines mark the 10% and 30% decision thresholds). The five lesion-level classes concentrate below 10%; mosaic virus disease sits alone at another order of magnitude and is clearly bimodal — the right peak corresponds to whole-leaf annotation, the left peak to the lesion-level boxes mixed in.

The dataset also contains images composited from several photographs. In one instance, a leaf photograph in the upper-left region of a single image is labelled mosaic virus disease (15.4% of the image) while the berry region on the right carries nine small grape powdery mildew boxes — **two granularities within one image**, a direct manifestation of a non-uniform annotation standard.

One over-generalisation must be avoided here: the claim of this paper is *not* that a larger box yields a higher AP. Grape botrytis cinerea ranks second in box area yet has the lowest AP of the six classes (Table 7). The claim is confined to the specific shortcut formed by mosaic virus disease, together with the one global effect common to coarsely annotated classes that the placebo control of Section 4.4 reveals.

The negative control, and the earlier pipeline validation, use the FieldPlant dataset[10] (Roboflow Universe, workspace `plant-disease-detection`, project `fieldplant`, version 11, CC BY 4.0; 5156 images, 27 classes, covering cassava, maize and tomato). Related observations appear in Section 5.2 and the use as a negative control in Section 4.2.

### 2.2 Experimental setup

All experiments follow a single-variable principle: the same dataset and split, with identical epochs, effective batch size, optimiser policy and hardware.

| Item | Setting | Note |
|---|---|---|
| Epochs | 50 | E5 in fact stopped at epoch 16; see Section 5.3 |
| Effective batch | 8 | E5 via 4×2 gradient accumulation; E4 is 4, not aligned |
| Optimiser | auto / AdamW | Learning rate determined by the framework |
| Mixed precision | on | AMP for YOLO, bfloat16 for RF-DETR |
| GPU | RTX 4060 Laptop 8 GB | Consumer-grade laptop GPU |
| Framework | PyTorch 2.13.0+cu132 | Ultralytics 8.4.106[15] / RF-DETR 1.8.3[16] |
| Random seed | 0, deterministic | |

The effective batch size of E4 (RT-DETR-l[17]) is 4, inconsistent with the other groups, so it is not a strict single-variable control. We report it faithfully nonetheless: with 32 M parameters and a different detection paradigm it finishes level with the others, which supports the judgement of this paper regarding the room left for model-side optimisation.

Training of E5 stopped at epoch 16, manifesting as a data-loading rate unable to keep pace with GPU computation. The bottleneck was not localised further at the time, and we cannot distinguish whether it came from disk reads or from CPU-side preprocessing. The problem is a limitation of the engineering environment and unrelated to the model itself; Section 5.3 gives a verification that the weights used are representative.

### 2.3 Evaluation protocol

All metrics are reported on the test set, which was not used at any point during training or tuning; the validation set is used only for weight selection. This distinction has practical consequence: validation mAP50 for E1 is 0.872 against a test value of 0.8357, a difference of 3.6 percentage points.

Metrics are computed uniformly by pycocotools (the official COCO evaluation implementation[14]) at a confidence threshold of 0.01. This protocol must be held fixed — for the same E1 weights, the built-in Ultralytics evaluator returns 0.845 while pycocotools returns 0.8357, a difference of 0.9 percentage points, of the same order as the between-group differences at issue in this paper. Comparisons that mix the two protocols do not hold.

## 3 Performance ceiling and bottleneck

### 3.1 Five controls

**Table 3 Test-set results (single pycocotools protocol)**

| ID | Model | Params | Resolution | Batch | mAP50 | mAP50-95 |
|---|---|---|---|---|---|---|
| E1 | YOLO11n | 2.58 M | 640 | 8 | 0.8357 | 0.5159 |
| E2 | YOLO11s | 9.42 M | 640 | 8 | 0.8493 | 0.5168 |
| E3 | YOLO11n | 2.58 M | 960 | 8 | 0.8459 | 0.5130 |
| E4 | RT-DETR-l | 32 M | 640 | 4 | 0.8427 | 0.5220 |
| E5 | RF-DETR-Nano | 30.2 M | 384 | 8 | 0.8511 | 0.5230 |

The range of mAP50 is 1.54 percentage points and that of mAP50-95 is 1.00. This interval covers a 12.4-fold span in parameter count, a 6.25-fold span in pixel area, and one change of detection paradigm.

That range of 1.54 percentage points is insufficient to support a ranking of the groups, on two grounds.

The first is indirect: E2 and E3 both exceed E1 on the test set but both fall below E1 on the validation set — opposite directions.

The second is direct: the E1 configuration (YOLO11n@640) was retrained with four further random seeds, every other setting held fixed, giving five runs in total including seed 0, evaluated through the same pipeline:

**Table 4 Five-seed repetition of one configuration (YOLO11n@640, n = 5)**

| Metric | Mean | SD |
|---|---|---|
| mAP50 | 0.8249 | 0.0069 |
| mAP50-95 | 0.5076 | 0.0063 |
| AP_small | 0.1663 | 0.0080 |
| AP_medium | 0.3661 | 0.0245 |
| AP_large | 0.5865 | 0.0080 |

The standard deviation of mAP50 for a single configuration is 0.69 percentage points. The 1.54-point range across the five configurations in Table 3 amounts to about 2.2 standard deviations — and the expected range of five draws from a single distribution is itself around 2.3 standard deviations. The observed between-group differences are entirely explicable by the random seed.

Two supplementary remarks.

First, each group in Table 3 reports a single run at seed 0. For E1, seed 0 happens to be the highest of the five (0.8357), 1.08 percentage points above the five-seed mean — a deviation of the same magnitude as the 1.54-point between-group range. Using single runs for between-group comparison is unreliable at this data scale.

Second, the number of repetitions itself must be sufficient. Taking AP_medium as an example, the first three seeds give 0.3788/0.3820/0.3847 with a standard deviation of only 0.0030, apparently extremely stable; extending to five seeds (adding 0.3263 and 0.3588) raises the standard deviation to 0.0245, eight times the original. A variance estimate at n = 3 can badly understate the true variation, and significance judged on that basis readily yields wrong conclusions.

The seed-to-seed variation of the **per-class** AP for the same set of weights was computed separately, to serve as the noise reference for the counterfactual comparison in Section 4.4; the values appear in the last column of Table 10. The relative standard deviation lies between 1.0% and 5.6% across classes, largest for grape botrytis cinerea (117 test instances, the fewest), whose relative range across the five runs reaches 15.5%. **Seed-to-seed variation of per-class metrics is appreciably larger than that of aggregate metrics, and any single-run per-class comparison must take this as its noise floor.**

## 3.2 By scale: the bottleneck is small objects

**Table 5 Scale-stratified performance of the five experiments**

| ID | AP_small | AP_medium | AP_large | large/small |
|---|---|---|---|---|
| E1 | 0.1567 | 0.3788 | 0.5967 | 3.81× |
| E2 | 0.1642 | 0.4275 | 0.5847 | 3.56× |
| E3 | 0.1856 | 0.4520 | 0.5852 | 3.15× |
| E4 | 0.1927 | 0.4357 | 0.5969 | 3.10× |
| E5 | 0.1614 | 0.3282 | 0.5982 | 3.71× |

AP on large objects is three to four times that on small objects, without exception across the five architectures. This indicates that what limits performance is not the model's capacity to represent the visual features of disease, but the scale of the targets themselves; changing architecture does not remove the gap. The size bands follow the COCO definition (small = area below 32×32 pixels)[14].

## 3.3 The effect of resolution is masked by aggregate metrics

E3 differs from E1 only in training and inference resolution (960 against 640). Judged by aggregate metrics the group yields no benefit whatever; but after repeating both groups over five seeds, the scale-stratified picture is entirely unlike the one the aggregate metrics give.

**Table 6 Scale-stratified comparison of E1@640 and E3@960 (five seeds each, Welch's t-test)**

| Metric | E1 (n=5) | E3 (n=5) | Difference | t | p |
|---|---|---|---|---|---|
| mAP50 | 0.8249 ± 0.0069 | 0.8370 ± 0.0096 | +1.20 pp | 2.28 | 0.055 |
| mAP50-95 | 0.5076 ± 0.0063 | 0.5055 ± 0.0116 | −0.21 pp | −0.36 | 0.731 |
| **AP_small** | 0.1663 ± 0.0080 | 0.1798 ± 0.0060 | **+1.35 pp** | 3.02 | **0.018** |
| AP_medium | 0.3661 ± 0.0245 | 0.4183 ± 0.0483 | +5.22 pp | 2.15 | 0.075 |
| AP_large | 0.5865 ± 0.0080 | 0.5809 ± 0.0103 | −0.56 pp | −0.96 | 0.367 |

**The improvement in small-object AP reaches statistical significance (p = 0.018), while aggregate mAP50-95 does not change at all (−0.21 percentage points, p = 0.731).**

A remark on multiple comparisons is required. Table 6 reports five metrics simultaneously, and judging all of them at $\alpha = 0.05$ raises a multiple-comparison problem. However, **AP_small is the pre-specified primary endpoint of this paper**: Section 3.2 independently identified small objects as the performance bottleneck across five architectures, and the present section was designed precisely to test whether raising resolution relieves that bottleneck — a hypothesis fixed before the experiment was run. The other four are secondary and reported for reference only; no separate conclusion is drawn from them.

A remark on the number of repetitions is likewise required. Section 3.1 notes that n = 3 understates variance, and the same objection applies in principle to n = 5. For the conclusion actually claimed here: the standard deviation of AP_small moves from 0.0094 at n = 3 to 0.0080 at n = 5, comparatively stable, whereas AP_medium moves from 0.0030 to 0.0245, a violent change. That this paper draws a conclusion only about the former and reports the latter merely as a trend rests exactly on that difference.

This is the principal observation of this paper concerning evaluation practice: the aggregate metric masks a genuine improvement in the bottleneck region. Judging by mAP50-95 alone one would conclude that raising resolution is ineffective, whereas in fact it improves precisely the bottleneck identified in Section 3.2 — small-object detection.

The effect size for medium objects is larger (+5.22 percentage points, about four times that for small objects) yet does not reach significance (p = 0.075), because the seed-to-seed variation of that metric is itself large (SD 0.0245 for E1 and 0.0483 for E3, three and eight times the small-object SD respectively). **AP_medium is an unstable metric on this data**, a point to bear in mind when designing further experiments — it requires more repetitions to support a conclusion of equal strength.

## 3.4 Scale dependence of recall

Average recall for E5 rises stepwise with the number of detections permitted: 0.2676 at maxDets 1, 0.5874 at 10, and 0.6729 at 100, not yet saturated. RF-DETR uses a fixed number of learned queries (300 in this experiment), and output is truncated when the number of objects in an image exceeds the number of queries. A single leaf in a real setting may carry dozens of lesions, and this limitation must be taken into account at deployment.

## 3.5 Class-level analysis

**Table 7 AP50-95 by class**

| Class | Test instances | E1 | E2 | E3 | E4 | E5 |
|---|---|---|---|---|---|---|
| grape black rot | 446 | 0.5668 | 0.5742 | 0.5620 | 0.5764 | 0.5649 |
| grape botrytis cinerea | 117 | 0.3454 | 0.3403 | 0.3167 | 0.3839 | 0.3801 |
| grape downy mildew | 170 | 0.4962 | 0.4925 | 0.5077 | 0.4579 | 0.4962 |
| grape powdery mildew | 313 | 0.3711 | 0.3816 | 0.3870 | 0.3892 | 0.4005 |
| grape ulcer disease | 133 | 0.5549 | 0.5689 | 0.5511 | 0.5609 | 0.5061 |
| mosaic virus disease | 63 | 0.7609 | 0.7435 | 0.7537 | 0.7636 | 0.7903 |

Computing the confusion matrix at the common operating point of confidence 0.25 and IoU 0.5 gives a result contrary to expectation: inter-class confusion barely exists (off-diagonal entries total only 4), and errors come almost entirely from missed detections and background false positives.

**Table 8 Detection outcome for E1 (conf ≥ 0.25, IoU ≥ 0.5)**

| Class | Instances | Correctly detected | Misclassified | Missed | Background FP | Recall |
|---|---|---|---|---|---|---|
| grape black rot | 446 | 420 | 0 | 26 | 36 | 94% |
| grape botrytis cinerea | 117 | 83 | 0 | 34 | 57 | 71% |
| grape downy mildew | 170 | 144 | 0 | 26 | 54 | 85% |
| grape powdery mildew | 313 | 241 | 0 | 72 | 157 | 77% |
| grape ulcer disease | 133 | 113 | 4 | 16 | 39 | 85% |
| mosaic virus disease | 63 | 58 | 0 | 5 | 16 | 92% |
| **Total** | **1242** | **1059** | **4** | **179** | **359** | **85%** |

Of all 1242 ground-truth instances, only 4 are assigned to another class (3 grape ulcer disease predicted as grape botrytis cinerea and 1 as grape powdery mildew), 0.32%; against 179 missed detections and 359 background false positives.

Grape botrytis cinerea has the lowest AP (0.3454), and the reason is not that it is hard to tell apart from grape ulcer disease — only 3 instances are confused between them — but that its recall is only 71%, with nearly a third of instances undetected, accompanied by 57 background false positives. Grape powdery mildew has the most background false positives (157) and its AP is also low (0.3711).

**The dominant error source on this data is localisation, not classification.** The model fails at "can the target be found", not at "once found, which class is it". This is consistent with the scale-stratified conclusion of Section 3.2: the bottleneck lies in detecting small-scale targets, not in the separability of the disease classes.

It should be noted that we earlier judged, from the normalised confusion-matrix plot produced by Ultralytics, that grape botrytis cinerea and grape ulcer disease were severely confused with each other. Re-checking against raw counts, that judgement does not hold, and it is corrected here.

One anomaly appears at this point: mosaic virus disease has the fewest test instances (63), yet its AP ranks first across all five architectures. The class with the smallest sample achieving the highest score runs contrary to the usual pattern; taken together with the difference in annotation area shown in Table 2, the two clues point to the same place.

# 4 Shortcut learning

## 4.1 Hypothesis

Two observations point at the same class: its annotation boxes are an order of magnitude larger than those of the other classes (Table 2); and it has the smallest sample yet the highest AP (Table 7).

From this we form the hypothesis that the model's high score on this class does not derive from recognising the symptoms of mosaic virus, but that the difference in annotation granularity supplies an exploitable shortcut.

## 4.2 Cross-species negative control

Testing this hypothesis requires a body of images that can be established as containing no grape mosaic virus. The 10 self-collected phone photographs are inadequate for the purpose: the sample is far too small, and whether a given leaf is infected with mosaic virus can only be judged by eye, which is not objective ground truth.

We use instead the 5156 images of FieldPlant. Its subjects are cassava, maize and tomato, which cannot contain grape mosaic virus, so any grape disease the model outputs on them is a false positive — ground truth that is objective and requires no human adjudication. Evaluation uses the E1 weights at a confidence threshold of 0.25.

**Table 9 Class distribution of false positives (full-resolution decoding)**

| Class | Training box area | Training share | Predicted share | Over-representation |
|---|---|---|---|---|
| mosaic virus disease | 43.16% | 4.9% | 65.7% | **13.41×** |
| grape botrytis cinerea | 8.12% | 12.9% | 31.3% | 2.42× |
| grape downy mildew | 2.39% | 14.5% | 2.4% | 0.16× |
| grape black rot | 0.57% | 35.3% | 0.5% | 0.02× |
| grape powdery mildew | 2.01% | 19.4% | 0.0% | 0.00× |
| grape ulcer disease | 3.29% | 12.9% | 0.0% | 0.00× |

Of the 5156 images, 3697 (71.7%) produced output, for a total of 12221 false-positive boxes. Among them, 1408 images (27.3%) were assigned grape mosaic virus at a confidence above 0.5, 744 above 0.7, with a maximum of 0.948.

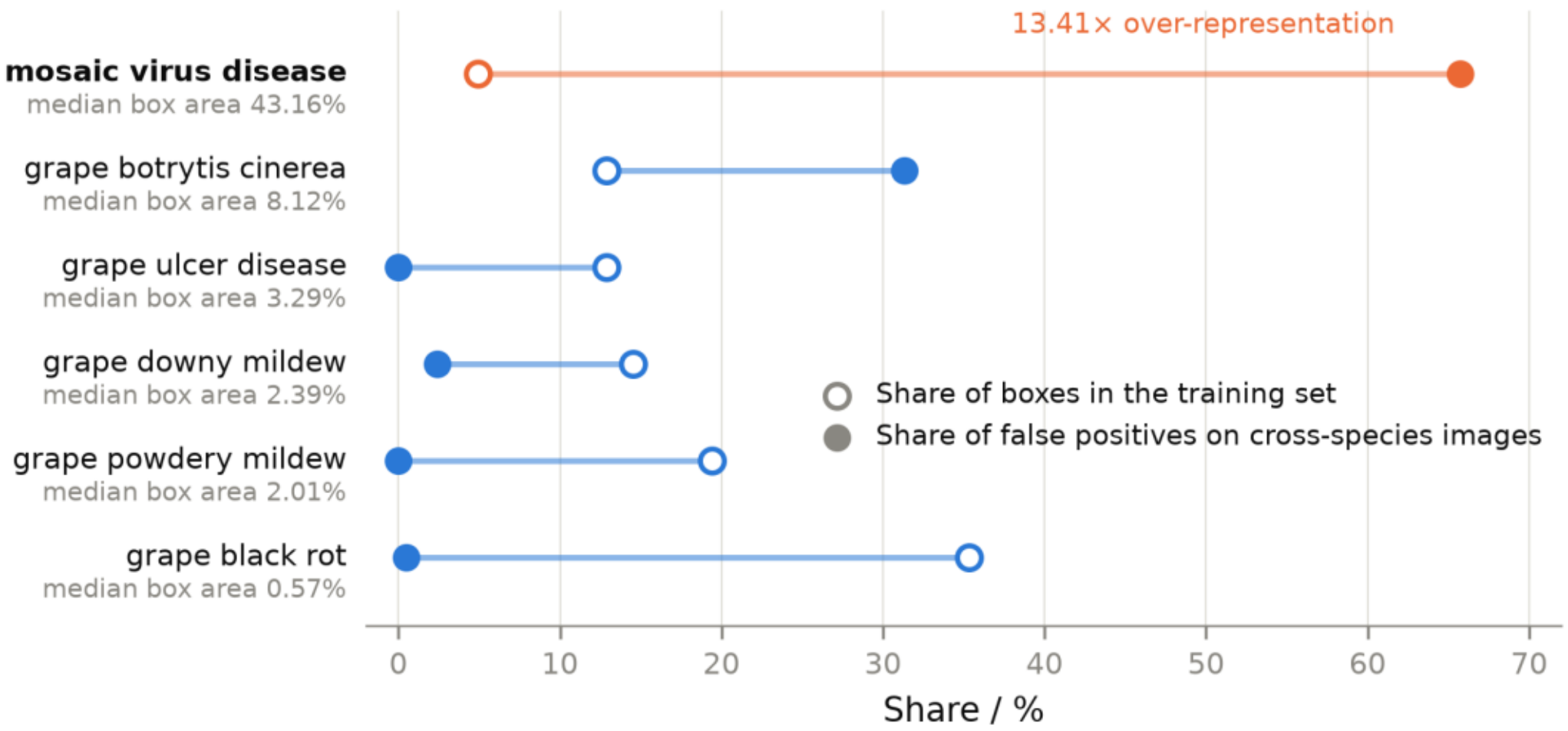

**Figure 2** Share of annotation boxes in the training set (hollow) against share of false positives on cross-species negative images (filled), for each class, ordered top to bottom by decreasing median training box area. Every class other than mosaic virus disease has a false-positive share below its training share; that one class — the most coarsely annotated and the least frequent in training — moves sharply in the opposite direction, from 4.9% to 65.7%.

**One necessary exclusion.** The most over-represented class happens to be the rarest in the training set (4.9%). A class-frequency prior would only cause high-frequency classes to be over-predicted, the direction opposite to what is observed (Spearman correlation −0.638). The bias therefore cannot be attributed to class imbalance.

**Two confounders, distinguished.** It is necessary to separate "why false positives occur at all" from "why false positives concentrate in one class"; these are different quantities, and the claim of this paper concerns only the latter.

First, the training set contains almost no purely healthy images without disease signs, so the model has never seen the state "no disease sign". This suffices to explain the overall *level* of false positives (71.7% of images produced output), but not their class distribution — the absence of healthy negatives has no directional effect across the six classes and cannot explain why 65.7% of the false positives concentrate in the class with the smallest training share. More importantly, this factor is held constant across the three counterfactual groups of Section 4.4: the three use the same training images, equally devoid of healthy samples, and the sole difference between them is the size of the annotation boxes. It therefore affects the **level** of false positives and not the between-group **difference**, and the causal claim of this paper rests on the latter. The independent effect of healthy negatives requires a separate experiment to measure and is listed as future work (Section 6).

Second, FieldPlant differs from the main dataset in acquisition conditions as well as in crop species, so the total number of false positives contains a contribution from domain shift. This factor likewise acts on the level and not on the class distribution, and the three counterfactual groups of Section 4.4 are compared on the same set of negative images, so domain shift is common to all three.

## 4.3 An explanation that was not supported

We originally attributed the phenomenon to a specific rule: that the model predicts mosaic virus whenever a large, leaf-shaped green region is present in the frame. The rule is superficially plausible but does not survive testing.

Computing, for each image, the fraction occupied by the largest connected green region, and comparing it against the rate at which the class is predicted:

| Largest green region fraction | Images | Rate of predicting this class |
|---|---|---|
| 0.00–0.05 | 31 | 6.5% |
| 0.05–0.15 | 171 | 14.6% |
| 0.15–0.30 | 678 | 40.7% |
| 0.30–0.50 | 1222 | 52.6% |
| 0.50–0.70 | 1230 | 51.7% |
| 0.70–1.01 | 1824 | 30.4% |

The binned rates form an inverted U, but the overall correlation is only −0.092, effectively zero; the median green fraction is 0.570 for images with no output and 0.586 for images with output, indistinguishable. Moreover the median area of the predicted boxes for this class is only 3.8%, and does not reproduce the 43.2% of the training annotations.

**This explanation is withdrawn.** The phenomenon itself is robust, but the specific image feature the model relies on remains unknown.

## 4.4 Counterfactual retraining

With the triggering feature unlocatable at the image level, we manipulate the variable directly at the data end instead.

Two counterfactual datasets were constructed, with training hyperparameters identical to E1 in every respect:

The **shrink** group retains the class and all of its samples, cropping every annotation box to the central 25% of its area; the median area fraction falls from 43.16% to 10.79%, the same order as grape botrytis cinerea. Images, box count, class count and class identity are all unchanged; the only thing changed is annotation granularity.

The precondition for this operation is that the symptom of mosaic virus is a whole-leaf diffuse mottling rather than a local lesion, so that after cropping to the centre the box still contains genuine symptomatic tissue. The precondition does not hold for local round lesions such as those of black rot, and the operation is therefore applied to this class only.

It should be noted that about 28% of this class's boxes were already lesion-level (area below 10%, see Section 2.1), and applying the same crop to them shrinks them further. We did not treat the two parts differently, in order to keep the operation single-variable. The consequences differ in direction for the two kinds of metric and must be stated separately.

For **cross-species false positives**, only 58% of the class actually undergoes the whole-leaf → lesion-level conversion; cropping the remainder is irrelevant to the hypothesis of this section and amounts to diluting the treatment. The measured −66% is therefore a **conservative estimate** of the effect.

For **in-distribution AP**, further cropping boxes that were already small adds detection difficulty and may inflate the apparent decline in AP. This point shares a root with the principal threat to validity discussed below and should be considered together with it.

The **drop** group removes the class outright. Training images in this group fall from 2631 to 2354 (a reduction of 10.5%, because some images contain only annotations of this class), which introduces a sample-size confound, so it serves only as corroborating evidence.

The criterion was fixed before the experiment was run: if the over-representation of the shrink group on FieldPlant falls back appreciably, then annotation granularity has a causal role in the shortcut.

The cross-species tests in this section all use quarter-resolution decoding. FieldPlant consists of roughly 13-megapixel originals, full-resolution decoding costs 3.4 times as much time as quarter-resolution, and all images are in any case scaled to 640 before entering the model. The E1 baseline was re-measured under this setting, giving an over-representation of 11.95× (against 13.41× at full resolution). The three groups share the same setting and are internally comparable.

**Result 1: the anomalous in-distribution score disappears once granularity is normalised**

Each group is evaluated under its own annotation scheme — the test annotations of the shrink group are shrunk correspondingly — with metrics computed by pycocotools under the same protocol as Tables 3 and 7. The last column gives the relative range of per-class AP over five random seeds of the same configuration (Section 3.1), as a noise reference.

**Table 10 Per-class in-distribution AP50-95 before and after shrinking (single pycocotools protocol)**

| **Class** | **E1 original** | **Shrink group** | **Change** | **Seed-to-seed relative range (n=5)** |
|---|---|---|---|---|
| mosaic virus disease | 0.7609 | 0.4552 | **−40.2%** | 3.2% |
| grape black rot | 0.5668 | 0.5600 | −1.2% | 2.5% |
| grape botrytis cinerea | 0.3454 | 0.3184 | −7.8% | 15.5% |
| grape downy mildew | 0.4962 | 0.4785 | −3.6% | 8.2% |
| grape powdery mildew | 0.3711 | 0.3809 | +2.7% | 5.8% |
| grape ulcer disease | 0.5549 | 0.5503 | −0.8% | 2.1% |

For the five classes that were not manipulated, whose test annotations are byte-identical between the two datasets, the measured change in AP is in every case smaller than that class's own seed-to-seed variation — indistinguishable at the resolving power available here. The manipulated class falls by 40.2%, 12.6 times its seed-to-seed relative range (3.2%). It drops from a commanding first place (0.7609, against 0.5668 for the runner-up) to fourth of the six (0.4552). The anomaly of Table 7 — the smallest sample scoring highest — disappears once annotation granularity is normalised.

The choice of evaluation arrangement requires explanation. Evaluating the E1 weights directly against the shrunk test annotations gives an AP of 0 for this class: with box area cropped to 25%, the IoU of two concentric boxes is bounded above by 0.25, so the two annotation schemes barely overlap at IoU ≥ 0.5. Each group must therefore be evaluated under its own annotation scheme, and what is compared is the *rank of the class relative to the other five*, not absolute values across annotation schemes.

**Result 2: cross-species false positives fall sharply**

**Table 11 False positives of the counterfactual retrainings on FieldPlant (quarter-resolution decoding)**

| Group | FP boxes of this class | Over-representation | Total FP |
|---|---|---|---|
| E1, original annotations (box 43.2%) | 5379 | 11.95× | 9176 |
| Shrink group (10.8%) | 1805 (−66%) | 6.16× (−48%) | 5971 |
| Drop group | 0 | — | 7056 |

**Result 3: removing the class does not eliminate the problem**

**Table 12 Transfer of false positives after removing the target class (quarter-resolution decoding)**

| Class | Median training box area | E1 | Drop group |
|---|---|---|---|
| grape botrytis cinerea | 8.12% | 3537 | 5350 (+51%) |
| grape downy mildew | 2.39% | 215 | 1225 (+469%) |

False positives transfer to the next most coarsely annotated classes. This shows that the shortcut does not attach to the particular disease mosaic virus disease: once that class is removed, out-of-distribution input is taken up by the others instead. It also explains why the shrink group (5971 total false positives) outperforms the drop group (7056): the former changes the condition that produces the shortcut, while the latter merely removes the class that had been receiving it.

The classes that take up the load happen to be the next most coarsely annotated, which suggests that the coarsest class is simply where out-of-distribution input goes. Natural as it looks, that inference cannot be established from this group: botrytis and downy mildew were already the second and third largest recipients of false positives under E1 (3537 and 215, Table 9), so their order of succession predates the manipulation and is confounded with the granularity ranking. Establishing the inference requires an independent test in the opposite direction — coarsening a fine-grained class that was not a recipient, and asking whether it becomes one. Result 5 answers that question in the negative.

**Two reservations**

First, annotation granularity is not the whole cause. An over-representation of 6.16× remains after shrinking; granularity accounts for roughly half, and the origin of the remainder is unclear.

Second, the principal threat to validity is that shrinking the annotation boxes itself increases detection difficulty, so a decline in AP does not necessarily indicate a shortcut. There are two responses. On the one hand, an increase in task difficulty cannot explain a 66% reduction in cross-species false positives, which bears no direct relation to box size; the two results must be explained jointly for the account to hold. On the other hand, after normalisation the AP of this class is 0.4552, in the same band as grape downy mildew (0.4785) and grape botrytis cinerea (0.3184), whose boxes are of comparable area — targets at this scale are not undetectable, and what falls away is the class's **anomalous advantage** relative to the others, not its absolute detectability.

**Result 4: the effect is specific to the manipulated class**

To exclude the blanket explanation that "any perturbation of annotations reduces false positives", a third control was run: leaving mosaic untouched and instead shrinking the boxes of grape botrytis cinerea (second largest in training box area at 8.12%) to 25% of their area (reducing it to 2.03%), all other conditions unchanged.

**Table 13 False positives per image for the three counterfactual groups (quarter-resolution decoding)**

| Group | mosaic | botrytis | Total FP |
|---|---|---|---|
| E1, original annotations | 1.043 | 0.686 | 9176 |
| mosaic boxes shrunk | **0.350 (−66%)** | 0.710 (+4%) | 5971 |
| botrytis boxes shrunk | 0.579 (−44%) | **0.108 (−84%)** | 3956 |

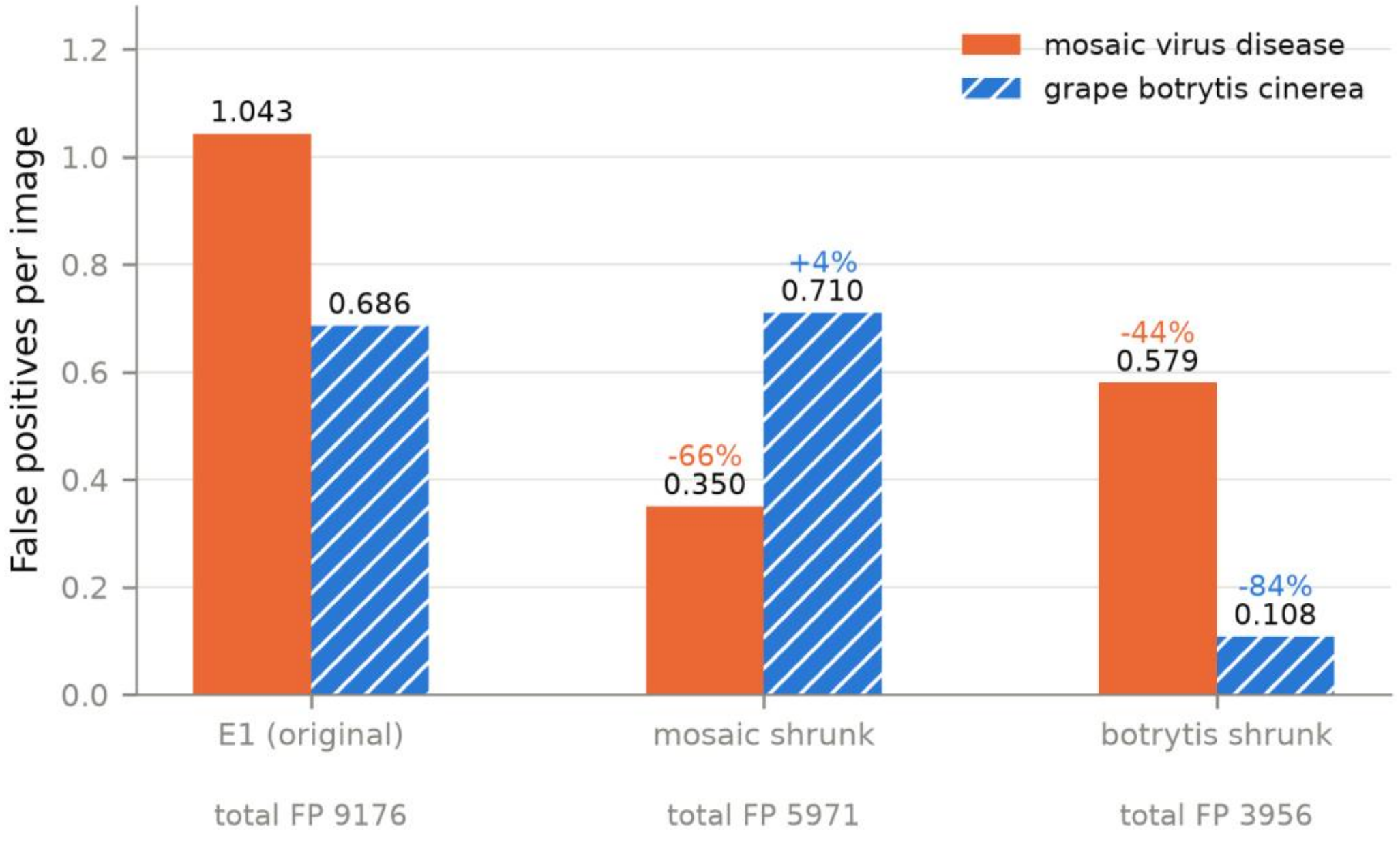

**Figure 3** False-positive boxes per image on cross-species negative images for the three counterfactual groups (quarter-resolution decoding); percentages are changes relative to E1 with original annotations, and the total false-positive count is given below each group. The middle group demonstrates the specificity of the effect (the manipulated mosaic falls by 66% while the unmanipulated botrytis does not move); the right-hand group demonstrates the difference in direction between the two mechanisms — shrinking botrytis lowers its own false positives by 84% while the unmanipulated mosaic also falls by 44%, the opposite of the rise that the sink mechanism acting alone would produce.

The effect concentrates on the manipulated class: shrinking mosaic lowers its false positives by 66% while botrytis barely moves (+4%); shrinking botrytis lowers its own by 84%. This excludes the explanation that perturbing annotations reduces false positives in general.

**Result 5: coarsening a fine-grained class does not create a new sink**

All three manipulations so far run from coarse to fine, and can therefore show only that weakening granularity weakens the shortcut, not that granularity suffices to produce one. If the sink is indeed determined by whichever position is annotated most coarsely, then the manipulation in the opposite direction should be equally effective: coarsen a fine-grained class to whole-leaf level and it should become the new destination for out-of-distribution input. This group tests that inference directly.

The class chosen is grape black rot. Its median box area is 0.57%, the finest of the six, and it accounts for just 0.41% (38/9176) of E1's cross-species false positives — it has never been a sink. It is a fungal disease presenting as discrete brown circular lesions, with no biological ground for a whole-leaf annotation of the kind a systemic infection might justify. The group therefore doubles as a test of a separate objection: whether the shortcut turns on annotation granularity or on the symptom morphology of the disease itself. If coarsening makes this class a sink too, the former holds.

The operation involves no tunable coefficient: within each image, all lesion boxes of this class are replaced by their single enclosing box — precisely the box an annotator working at whole-leaf granularity would draw. This makes the class match mosaic virus disease on several structural properties at once.

**Table 14 Structural comparison of coarsened grape black rot with mosaic virus disease**

| | **grape black rot (coarsened)** | **mosaic virus disease** |
|---|---|---|
| Annotation boxes | 476 | 593 |
| Share of all annotations | 5.8% | 4.9% |
| Boxes per image (median) | 1 | 1 |
| Median box area | 40.37% | 43.16% |
| In-distribution AP50-95 | 0.8547 | 0.7338 |
| **Cross-species false-positive boxes** | **0 (0.0%)** | **3718 (50.0%)** |

The criteria were fixed before training completed, and the primary measure is the **raw predicted share** rather than the over-representation ratio: coarsening necessarily reduces the box count (whole-leaf annotation is one box per leaf by construction), taking this class from 4237 boxes to 476 and its share of training annotations from 35.3% to 5.8%, so a denominator 6.1 times smaller would inflate the ratio mechanically. The criteria have three bands — a predicted share above 10% together with a marked fall in the original class means the destination follows granularity; between 2% and 10% means granularity produces only a partial tendency; below 2% with the original class unmoved means the inference fails.

The measurement falls in the third band: the class emits **zero boxes** across 5156 cross-species images. This is not an artefact of the confidence threshold. Lowering the threshold from 0.25 to 0.001 and re-measuring (every sixth image, 860 in total) gives 207 boxes with a maximum confidence of 0.630 under E1, against 33 boxes with a maximum confidence of 0.009 here. The same model achieves an in-distribution mAP50 of 0.990 on this class, the highest of the six — its ability to detect the class is intact; it simply never assigns unfamiliar input to it.

**The fall in box count does not account for this result.** The class does lose 89% of its boxes. But mosaic virus disease is itself a class of only 593 boxes, 4.9% of all annotations, almost identical to coarsened grape black rot (476 boxes, 5.8%) — and it receives half of all false positives. On this dataset, of two classes that are both coarse and rare, one is the largest sink and the other receives nothing at all.

**A further observation: the unmanipulated mosaic also falls.** Only the annotations of grape black rot were altered in this group, yet the cross-species false positives of mosaic virus disease fall from 5379 boxes to 3718 (−31%), while the remaining four classes together go from 3759 boxes to 3714 (−1.2%) — essentially unmoved.

This is not a global rescaling caused by the drop in total annotations. Coarsening takes the dataset from 11995 boxes to 8234 (−31%); had the model's propensity to emit output been scaled down accordingly, the other four classes would have fallen as well, whereas the measurement is −1.2%. The fall is specific to mosaic. The pattern runs in the same direction as the placebo group: manipulating a class *other than* mosaic lowers the unmanipulated mosaic in both cases (−44% in the placebo group, −31% here), and since the two groups manipulate different classes these are two independent observations. The patterns are not identical — the placebo group has the lowest total false positives of the three (3956), while here the other four classes do not move at all — but both show a cross-class effect, whose mechanism is not further resolved here.

**Conclusion: granularity determines magnitude, not destination**

The four counterfactual groups — shrink, drop, placebo and reverse manipulation — must be read together. The shrink and placebo groups show that changing a class's annotation granularity changes both that class's own out-of-distribution false positives (−66%, −84%) and the total number of false positives (9176 → 5971 → 3956); the cross-class effect is observed twice independently — shrinking the boxes of botrytis makes **the unmanipulated mosaic fall by 44%**, and coarsening black rot makes it fall by 31% (while the other four classes move by only −1.2%) — so altering any other class reduces not only that class's own share. In that respect annotation granularity has a causal effect on the **magnitude** of the shortcut, and the benefit is not confined to the class that is corrected, so it is not appropriate to correct only the coarsest class — all coarsely annotated classes should be treated together.

The reverse manipulation shows, however, that coarsening an annotation does not make a class the destination for out-of-distribution input: with granularity, box count and share of annotations all matched to mosaic virus disease, and with in-distribution performance if anything higher (AP50-95 0.8547 against 0.7338), grape black rot receives zero cross-species false positives.

The hypothesis of Section 4.1 must therefore be narrowed: **annotation granularity is a modulator of this shortcut, not its cause.** It can amplify or attenuate a sink that already exists, but is not sufficient to create one. The transfer of false positives to the next-coarsest classes in the drop group should correspondingly be read as a redistribution among pre-existing tendencies to receive, not as granularity conferring the capacity to receive.

**What determines the destination is not established here.** Between coarsened grape black rot and mosaic virus disease, every annotation property measured has been matched, so the difference between them can only lie in the content of the images themselves. One untested candidate explanation is that out-of-distribution input is assigned to the class whose training images are closest to it in overall appearance, rather than to the class with the largest boxes. Testing that explanation requires a distributional distance measured at the image level, which is not attempted here and is listed as future work (Section 5.3). The group also shows that the gap left when the green-region hypothesis was withdrawn in Section 4.3 — what the model actually uses to assign unfamiliar input to a class — has not been filled by annotation granularity.

### 4.5 Why this failure mode is particularly dangerous

This failure mode does not present as "failure to detect" but as "stably emitting the same wrong answer", and its danger is threefold.

**High confidence, unresponsive to thresholding.** Raising the threshold from 0.25 to 0.5 does not make the misclassification disappear: 1408 of the 5156 negative images are still assigned mosaic virus at a confidence above 0.5, and 744 above 0.7. Raising the threshold merely loses true positives as well.

**Invisible to routine monitoring.** The output format is normal and the confidence distribution is plausible; without per-sample visual inspection the problem cannot be noticed.

**Its cost exceeds that of no output at all.** If such a system were put into service, the user would spray on the basis of a wrong conclusion — a cost higher than that of a system reporting plainly that it cannot decide.

## 5 Discussion

### 5.1 The feasibility boundary of airborne lesion-level detection

Our initial reasoning was: small-object AP is only 0.16, so the ground sampling distance must be reduced, so a two-tier flight plan is needed. That reasoning has a flaw — the value 0.16 was measured on ground-level close-range images, and whether it extrapolates to an aerial viewpoint had never been verified.

We compute instead directly from imaging optics, without relying on any cross-domain assumption:

GSD = pixel pitch × flight altitude / focal length

The COCO size bands are defined on the image and are independent of how the image was captured (small = area below 32×32 pixels)[14].

**Table 15 Resolving power of several cameras at 30 m altitude**

| Camera | GSD | Target size needed to reach 32 px |
|---|---|---|
| 1/1.3″ 48 MP, f = 6.7 mm | 5.33 mm/px | 17.1 cm |
| 4/3 20 MP, f = 12.29 mm | 8.00 mm/px | 25.6 cm |
| 1″ 20 MP, f = 8.8 mm | 8.22 mm/px | 26.3 cm |
| 1/1.3″ 12 MP telephoto, f = 43 mm | 1.66 mm/px | 5.3 cm |

**Table 16 Maximum altitude for a 5 mm lesion to reach a given pixel size (1/1.3″ 48 MP)**

| Requirement | Maximum altitude | Single-frame coverage | Frames per hectare |
|---|---|---|---|
| 32 px (leaving the small band) | 0.88 m | 1.26 × 0.94 m | ≈ 140 000 |
| 96 px (entering the large band) | 0.29 m | 0.42 × 0.31 m | — |

Frame counts assume the photogrammetric convention of 80% forward and 70% side overlap. Even in the idealised case of no overlap, covering one hectare still requires about 8400 frames; the conclusion is unaffected by that assumption.

At 30 m altitude, a 5 mm lesion occupies just 0.94 pixels.

**The conclusion is unambiguous: airborne detection of millimetre-scale lesions cannot be achieved at any operationally feasible altitude.** This is an optical limitation rather than a question of model capability — when a target occupies less than one pixel on the sensor, the information is already lost at acquisition and no small-object detection method has anything to work with. A telephoto lens can push the minimum resolvable size down to 5.3 cm, the most effective single measure available, but that is still an order of magnitude away from the millimetre scale.

What is required is therefore not a two-tier *flight plan* but a two-tier *system*: the airborne tier (30–50 m) takes on anomaly localisation at canopy and whole-plant scale — discoloured regions, abnormal vigour, missing plants, all of which exceed 17 cm and are optically resolvable — while lesion-level confirmation is carried out on the ground at close range. The dataset and models used in this paper correspond to the latter tier.

It must be emphasised that the altitudes above are a necessary condition only. The detection accuracy actually attainable at such altitudes must be measured on real aerial data, whereas all experiments in this paper use ground-level close-range images.

## 5.2 Data construction

**We first give a screening statistic that can be applied before the data is used.** The phenomenon of Section 4 originates in the inconsistency of annotation granularity across classes, and that inconsistency can be computed directly without reading a single image or training a single model. Let *m*<sub>c</sub> be the median relative area of the annotation boxes of class *c* (under normalised YOLO coordinates the relative area is simply *w*×*h*), and let

*R* = max<sub>c</sub>(*m*<sub>c</sub>) / median<sub>c</sub>(*m*<sub>c</sub>)

characterise how far the coarsest class departs from a typical class. The denominator is the median across classes rather than the minimum, so that a single extremely fine class cannot inflate it. The decision thresholds are set a priori: *m*<sub>c</sub> ≥ 30% is recorded as whole-leaf level (a box covering more than three tenths of the image cannot enclose a single lesion), *m*<sub>c</sub> ≤ 10% as lesion level, with no determination made in between.

*R* characterises *how uneven* the annotation is, but does not suffice to characterise *whether the coarsest class stands alone*. The latter is given by the ratio of the median of the coarsest class to that of the second coarsest, *m*<sub>(1)</sub>/*m*<sub>(2)</sub>, and the two must be read together: a dataset can be coarse overall with no gap at the head of the ranking, and it can equally be fine overall while containing one isolated coarsest class. Both quantities characterise how internally consistent the annotation scheme is, not how the model behaves; as shown below, an *m*<sub>(1)</sub>/*m*<sub>(2)</sub> close to 1 is not sufficient to conclude that a dataset contains no sink.

**Table 17 Annotation granularity of two datasets under a single protocol**

| **Dataset** | **Classes** | **Boxes** | **Median *m* across classes** | **Coarsest *m*** | ***m*<sub>(1)</sub>/*m*<sub>(2)</sub>** | **Whole-leaf/lesion classes** | **Granularity spread *R*** |
|---|---|---|---|---|---|---|---|
| Main dataset (grape) | 6 | 11995 | 2.84% | 43.16% | **5.32×** | 1/5 | **15.2×** |
| FieldPlant (cassava/maize/tomato) | 27 | 8580 | 33.49% | 87.46% | 1.06× | 17/1 | 2.6× |

The two indices differ in their sensitivity to small classes. Restricting the classes entering the statistic to those with at least 10, 30 and 100 boxes, the main dataset stays at 15.2× and 5.32× throughout (its smallest class still has 593 boxes, so thresholds have no effect); for FieldPlant, *R* remains stable between 2.4× and 2.7× while *m*<sub>(1)</sub>/*m*<sub>(2)</sub> rises from 1.06× to 1.89×, because removing small classes changes which class heads the ranking. Even at the top of that interval it remains far below the 5.32× of the main dataset.

**The two datasets fall into different regimes; "coarse" and "inconsistent" are two different things.** FieldPlant's annotations are at whole-leaf level throughout — 17 of its 27 classes have a median area above 30%, and the median across classes is itself 33.49% — but the classes lie close to one another: the coarsest exceeds the second coarsest by only 6% (1.06×), and there is no gap at the head of the ranking. The main dataset is the opposite: five classes are lesion-level (medians 0.57% to 8.12%) and one alone is whole-leaf level (43.16%), exceeding the across-class median by 15.2 times and the second coarsest class by 5.3 times (Section 2.1).

This contrast bears on the mechanism of Section 4, but the relation must be stated with care. The reverse manipulation of Section 4.4 (Result 5) is precisely a construction that drives *m*<sub>(1)</sub>/*m*<sub>(2)</sub> from 5.32× down to 1.07×: after coarsening, the main dataset has two whole-leaf classes side by side (43.16% and 40.37%), which by the screening statistics above places it in the "no gap at the head of the ranking" regime — the FieldPlant regime. If the sink were indeed determined by the coarsest position, the false positives should have split between the two classes. They did not: 50.0% remain concentrated in the original class and the newly coarsened class receives zero. *An m<sub>(1)</sub>/m<sub>(2)</sub> close to 1 therefore does not mean that a dataset contains no sink.*

This negative result bounds the scope of the two screening statistics without cancelling their value. *R* and *m*<sub>(1)</sub>/*m*<sub>(2)</sub> measure whether the annotation scheme is internally consistent; uneven granularity across classes is a defect that can be found before training and that has been shown to amplify out-of-distribution false positives (Section 4.4, Results 1 to 4). What they cannot serve as is a predictor of whether a dataset contains a shortcut sink. The claim of this paper is therefore stated as: **what is dangerous is not that annotation is coarse, but that granularity is uneven across classes — while even granularity is no guarantee that no shortcut exists**.

The practical implication is an executable screening step: compute *R* once before using a public detection dataset, at the cost of traversing the annotation files (seconds, with no images and no GPU required). The larger *R* is, the more the coarsest class should be inspected before training, or its annotation granularity unified.

The following recommendations all follow directly from the results above.

**Annotation granularity must be uniform across classes.** Section 4 shows that mixing whole-leaf and lesion-level annotation can give a model the highest in-distribution score on that class (0.74 to 0.79) while producing a 13.41-fold over-representation of false positives out of distribution; the problem cannot be detected from in-distribution metrics alone. The placebo control of Section 4.4 further shows that this requirement is not confined to the coarsest class: shrinking the boxes of any coarsely annotated class also lowers the total number of false positives, so all coarsely annotated classes should be regulated together rather than only the most conspicuous one.

**The training set should include a sufficient number of healthy (disease-free) samples.** The public dataset used here contains almost no purely healthy images, so the model has never seen the state "no disease sign", which may be one reason why 71.7% of images produced output in Section 4.2. This recommendation operates at a different level from the previous one: the absence of healthy samples affects **whether** the model emits output at all on unfamiliar input, whereas annotation granularity affects the **magnitude** of the false positives that do occur (Section 4.4). Granularity does not determine their class destination — that inference is refuted by the reverse manipulation of Section 4.4. The qualitative separation of the two factors is given in Section 4.2; the independent effect of the former has not been measured here and is listed as future work.

**Self-collected data should take part in training rather than serve only as a test set.** This recommendation is an inference from the preceding points together with recommendations in the literature; **this paper offers no direct experimental evidence for it** — Section 4.2 excluded the self-collected data from the formal experiments for insufficient sample size (10 images). It is listed to inform subsequent data construction, and its verification awaits the completion of annotation for the self-collected dataset.

**Public datasets must be checked by human visual sampling and not judged from metrics alone.** Sampling reveals more than mislabelling. In the main dataset used here, simply looking through the images reveals samples composited from several photographs, in one of which a leaf in the upper-left region is labelled mosaic virus disease while the berry region on the right carries nine small grape powdery mildew boxes — **two annotation granularities within one image** (Section 2.1).

Such problems appear in no metric, yet they are precisely the source of the shortcut described in Section 4.

During the earlier pipeline validation with FieldPlant, sampling the annotation images likewise revealed maize leaves labelled as a tomato class — maize is a monocot with parallel venation and differs obviously from the pinnate compound leaf of tomato, so this is a clear annotation error; there is also severe under-annotation, with a dozen or more leaves clearly visible in an image of which only 1 to 4 are annotated, causing genuinely present but unannotated targets to be counted as false positives when the model detects them. None of these problems can be found from metrics.

## 5.3 Limitations

**What determines the destination of the shortcut is not answered here.** This is the principal open question of the paper. The reverse manipulation of Section 4.4 shows that annotation granularity is not the answer: coarsening the finest class until it matches mosaic virus disease in granularity, box count and share of annotations — with in-distribution performance if anything higher (AP50-95 0.8547 against 0.7338) — still leaves its cross-species false positives at zero. The green-region hypothesis tested in Section 4.3 has likewise been rejected. The question of what the model actually uses to assign unfamiliar input to a class therefore remains open at the close of this paper.

One untested candidate explanation is that the sink is determined by how close a class's training images are to the out-of-distribution input in **overall appearance**, rather than by the size of its annotation boxes. That explanation is testable — for instance by measuring the distributional distance between each class's training images and the FieldPlant images in a feature space or through image statistics, and asking whether that distance predicts how many false positives each class receives. It is not measured here because it calls for a family of measures unlike those used elsewhere in this paper, and because its conclusion would depend on the chosen feature extractor, so it requires controls of its own. Were it to be supported, the phenomenon of Section 4 would be better restated as a problem of domain similarity, with annotation granularity demoted to a modulating factor on magnitude — consistent with the present conclusions, though the level at which the attribution is made would change.

**All experiments use ground-level close-range images; transfer to an aerial viewpoint is untested.** This paper makes no assertion about detection performance in UAV settings, and the conclusions of Section 5.1 rest on optics alone.

**The number of repetitions remains limited.** E1 and E3 were each run with five seeds; E2, E4 and E5 remain single runs. The between-group comparison of Table 3 can therefore only state that the differences are of the same order as seed variation, without discriminating further. Drawing a conclusion about any specific pair of groups requires repetitions of each in the manner of Section 3.1. The significance judgement for AP_small in Section 3.3 ($p = 0.018$) likewise rests on $n = 5$; although the variance estimate of that metric is comparatively stable between $n = 3$ and $n = 5$ (SD 0.0094 → 0.0080), larger-scale repetition remains necessary.

**The counterfactual groups are also single runs.** All three counterfactual groups (shrink, drop, placebo) were trained once at seed 0. To make single-run comparison interpretable, this paper uses the seed-to-seed variation of per-class AP over five seeds of the same configuration as a noise reference (last column of Table 10): the change in the manipulated class is 12.6 times its seed-to-seed relative range, far above the noise, while the changes in the five unmanipulated classes all fall within their respective noise ranges and can therefore only be judged as "no detectable change at the resolving power available here", not as a strict null effect.

**How common this regime is among public datasets is a question this paper cannot answer.** Section 5.2 measured annotation granularity on two datasets under a single protocol, of which only one exhibits the across-class inconsistency regime ($R$ = 15.2×) while the other is uniformly coarse ($R$ = 2.6×). Two samples cannot support any judgement about prevalence; answering that question requires a larger survey of public detection datasets. Moreover the causal chain of Section 4 was established on one dataset and its transfer to others is untested — in particular, whether the mechanism still holds on data with no isolated coarsest class is unevidenced.

**The absence of healthy negatives was not measured separately.** The main dataset contains almost no images without disease signs. This factor suffices to explain the overall level of false positives in the cross-species negative control but not their class distribution, and it is held constant across the three counterfactual groups (Section 4.2). The causal claim of this paper is therefore unaffected by it, but the question of how far adding healthy negatives would reduce the total number of false positives is one this paper does not answer.

**The number of epochs for E5 differs from the other groups.** That group stopped at epoch 16 owing to a data-loading constraint, and the best weights from epoch 13 were used. To confirm that those weights are representative, we resumed training from the interruption to epoch 29: validation mAP50 fluctuated within 0.849–0.867 over the subsequent 16 epochs with no upward trend, indicating that the model had converged by epoch 13 and that the weights used represent what this method achieves on this data. Its input resolution is fixed at 384 by the model architecture and cannot be aligned with the other groups.

**The effective batch size of E4 is 4**, inconsistent with the other groups, so it is not a strict single-variable control.

**The surface triggering feature of the shortcut remains unknown.** The phenomenon holds (Section 4.2) and the causal effect of annotation granularity on the magnitude of the shortcut holds (Section 4.4), but what image feature the model actually relies on has no answer; the hypothesis tested in Section 4.3 was refuted. Granularity also accounts for only about half, with 6.16× remaining after shrinking. What determines the destination is likewise undetermined; see above.

**The self-collected dataset is not yet annotated**, and quantitative evaluation of domain shift remains to be supplied.

**The design of the sliced-inference experiment (E7) was invalid.** We used SAHI[12] sliced inference to test the small-object hypothesis, and results fell across the board, mAP50-95 dropping from 0.516 to 0.292. Post-hoc analysis shows the design does not hold: the original images are 640, the same as the model input, so slicing and upscaling is pure upsampling that creates no new information while introducing scales not seen during training. The results support no conclusion; they are reported faithfully here to show that the criteria of this paper were in fact applied in practice.

**The field-scale spatial pattern analysis is not connected to real data.** The Clark-Evans nearest-neighbour index, Ripley's L function, Moran's I and Getis-Ord Gi* have been implemented and verified on synthetic data to distinguish four patterns, but the self-collected data were captured hand-held and carry no geographic coordinates, so this is not reported as a result and is left for future work once coordinate-bearing survey data are available.

# 6 Conclusions

The overarching question of this paper is which intrinsic properties of a public dataset determine whether a detection system trained on it is actually usable. The five conclusions below each answer one facet of that question, and they point in a common direction: under the data conditions studied here, what determines usability is a property of the data rather than the choice of model.

**The room for model-side optimisation is exhausted.** A 12.4-fold span in parameter count, a 6.25-fold span in pixel area and one change of detection paradigm give a test-set mAP50 range of only 1.54 percentage points under a single protocol. Further investment on the model side has very low marginal return.

**The bottleneck is small objects, consistently across five architectures.** The ratio of large- to small-object AP is stable between 3.10× and 3.81×. Raising the input resolution improves AP_small by 1.35 percentage points (+8.1% relative, $p = 0.018$) and AP_medium by 5.22 percentage points (+14.3% relative, not significant), but the improvement is masked by aggregate metrics — judging by mAP50-95 alone yields the opposite conclusion.

**The harm of coarse annotation is not confined to a single class.** Comparison of the three counterfactual groups shows that shrinking the boxes of any coarsely annotated class not only reduces that class's own out-of-distribution false positives sharply (−66% and −84%) but also lowers the total number of false positives (9176 → 5971 → 3956). Excessively coarse annotation therefore does not merely raise one class's false positives; it also raises the model's overall propensity to emit output on unfamiliar input. Unifying the annotation standard consequently has value beyond the individual class.

**Inconsistent annotation granularity within this public dataset induced quantifiable shortcut learning.** It produces a 13.41-fold over-representation of false positives on 5156 cross-species negative images, at high confidence, unresponsive to thresholding, and invisible to in-distribution evaluation. Counterfactual retraining establishes a causal effect of granularity on the **magnitude** of the shortcut: shrinking only that class's boxes lowers its in-distribution AP from 0.7609 to 0.4552 and reduces cross-species false positives by 66%, while removing the class outright transfers the false positives to the next most coarsely annotated classes.

**Annotation granularity does not, however, determine the destination of the shortcut.** A manipulation in the opposite direction, with criteria registered in advance, returns a negative result: coarsening the most finely annotated class (median area 0.57%) to whole-leaf level (40.37%), so that it matches the first class in box count and share of annotations with in-distribution performance if anything higher (AP50-95 0.8547 against 0.7338), leaves its false positives across 5156 cross-species images at zero boxes — with a maximum confidence of only 0.009 when the threshold is lowered to 0.001, against an in-distribution mAP50 of 0.990, the highest of the six classes — while the unmanipulated original class still holds 50.0% of all false positives. The claim of this paper is therefore that **annotation granularity is a modulator of this shortcut, not its cause**: it can amplify or attenuate a sink that already exists, but is not sufficient to create one. What determines the destination could not be established; the green-region hypothesis of Section 4.3 has been refuted, and the question remains open at the close of this paper (Section 5.3).

**Airborne detection of millimetre-scale lesions is optically unattainable at feasible flight altitudes.** The role of the UAV should be canopy-level anomaly localisation, with lesion-level confirmation carried out on the ground.

**Future work** (ordered by dependency, not by importance)

**First, build a self-collected dataset with uniform annotation granularity.** This is a precondition for everything else and responds directly to the finding of Section 4. Two points matter: all classes adopt the same granularity (lesion level); and a sufficient number of healthy samples is included. The latter also supplies a control this paper could not complete — adding healthy negatives while holding annotation granularity fixed would measure their independent contribution to the total number of false positives, separating quantitatively the two factors that Section 4.2 could only distinguish qualitatively.

**Second, determine what fixes the destination of the shortcut.** This is the main question the paper leaves open. The green-region hypothesis of Section 4.3 has been refuted, the reverse manipulation of Section 4.4 further rules out annotation granularity, and the 6.16× over-representation remaining after shrinking is still unexplained. The leading candidate is domain similarity — the sink being fixed by how close each class's training images are to the out-of-distribution input in overall appearance (Section 5.3). The test would be to measure the distributional distance between each class's training images and the FieldPlant images and ask whether it predicts how many false positives each class receives, supported by feature attribution or input ablation to identify whether the operative channel is texture, colour statistics or something else.

**Third, quantify the annotation cost of few-shot transfer.** On top of the dataset above, measure the minimum annotation volume required to add a new disease class or change crop. That figure determines whether the approach can be generalised — if adding one class requires thousands of boxes, it is unsustainable in deployment.

**Fourth, carry out canopy-level aerial validation under the optical constraints of Section 5.1.** Section 5.1 gives only a necessary condition (target size must exceed a given pixel count); the accuracy actually attainable must be measured on real aerial data. Note that this is not the same task as the lesion-level detection studied here and requires its own data and annotation standard.

**Fifth, time-series observation of disease development.** This depends on a fixed camera position: if viewpoint, distance and illumination differ between captures, results at different times are not comparable and time-series analysis loses its meaning. It is therefore best undertaken once the first and fourth items are in place, with observation points fixed in advance.

## References

Bibliographic details (authors, title, venue, volume, issue, pages, DOI) were verified item by item against the Crossref and arXiv APIs on 2026-08-13. References follow IEEE style. Entries [6], [7] and [9] were still arXiv preprints as of the verification date, with no record of formal publication. Where the full author list could not be verified, "et al." is used.

## Appendix Reproducibility

All experiments fix the random seed and are reproducible under identical software and hardware. The environment is Windows, RTX 4060 Laptop 8 GB, PyTorch 2.13.0+cu132, Python 3.11. Peak GPU memory use is 2.8 GB (YOLO11s, batch 8, resolution 640), far below the 8 GB limit — training and deployment of this approach do not depend on high-performance computing equipment.

All in-distribution metrics are computed by pycocotools on the same test set at the same confidence threshold, including the counterfactual groups of Section 4; per-class and scale-stratified metrics are all written to disk and can be checked in full.

| Script | Purpose | Output |
|---|---|---|
| `scripts/Eval_unified.py` | Single pycocotools protocol evaluation of the five models | `results/runs_unified/table3_unified.json` |
| `scripts/Eval_perclass_seeds.py` | Seed-to-seed variation of per-class AP (last column of Table 10) | `results/runs_unified/perclass_seed_variance.json` |
| `scripts/Shortcut_experiment.py` | Cross-species negative control | `results/runs_shortcut/shortcut_*.json` |
| `scripts/Counterfactual_prepare.py` | Construction of the counterfactual datasets | `datasets/grape_cf_*` |
| `scripts/Counterfactual_train.py` | Counterfactual retraining | `runs/detect/E10*`, `E12_placebo`, `E13_expand` |
| `scripts/Eval_counterfactual.py` | In-distribution evaluation of the counterfactual groups (Tables 10 and 14) | `results/runs_unified/counterfactual_unified*.json` |
| `scripts/Conf_sweep.py` | Low-threshold re-measurement separating a hard zero from a threshold effect (Result 5) | console output, see Section 4.4 |
| `scripts/Granularity_stats.py` | Cross-dataset annotation granularity statistics (Table 17) | `results/runs_unified/granularity_stats.json` |
| `scripts/Gsd_planner.py` | Acquisition parameters and feasibility calculation | Tables 15 and 16 |

The counterfactual datasets carry YOLO-format annotations only. When `scripts/Eval_counterfactual.py` constructs COCO annotations from them, image ids, file names and category ids are all inherited from the original test set and only the boxes are replaced. The correctness of this conversion is confirmed by a self-check: the five classes that were not manipulated give byte-identical AP under the old and new annotations.

The criteria for Result 5 of Section 4.4 were registered in advance: the three bands, the choice of primary measure and the reasoning behind it, together with the prediction derived from the screening statistics of that dataset, are recorded in the file header of `scripts/Shortcut_experiment.py`, written before that group finished training. The verdict is issued automatically by a 3×3 decision table in which all nine combinations carry a pre-assigned reading, so that no cell is left to be interpreted after the fact; the verdict and the screening statistics are written to disk alongside the false-positive counts, in the `prereg_verdict` and `screening` fields of `results/runs_shortcut/shortcut_E13.json`. Substituting the E1 baseline values into the table yields "refuted" — that is, "nothing changed" necessarily falls on the refuting side, so the table is not biased towards confirmation.

**Open resources and licences**

| Resource | Licence | Use |
|---|---|---|
| Grape disease detection dataset (`wscs/grape-uyimv` v1) | CC BY 4.0 | Main experiments |
| FieldPlant (`plant-disease-detection/fieldplant` v11) | CC BY 4.0 | Pipeline validation, cross-species negative control |
| Ultralytics YOLO11 | AGPL-3.0 | E1–E4 |
| RF-DETR | Apache 2.0 | E5 |
| pycocotools | BSD | Metric computation |
| SAHI | MIT | E7 |
| X-AnyLabeling | GPL-3.0 | Annotation of self-collected data |

Ultralytics is licensed under AGPL-3.0, which imposes open-source requirements on the distribution of derivative works; RF-DETR is Apache 2.0, with fewer commercial constraints. This difference has been taken into account in subsequent technology selection.

## Declaration on the use of generative AI

In the course of conducting this work and preparing this paper, the author used a generative AI assistant for the following purposes: organising the prose of the related-work section and the background material, language polishing, formatting of tables and layout, and writing the scripts used to check the experimental data and perform the statistical computations.

All research work was carried out independently by the author, including: formulation of the research question, experimental design and the setting of controls, data collection and processing, model training and evaluation, and the interpretation of results and formulation of conclusions. All experiments reported here were run on the author's own equipment, and the raw data, model weights and intermediate results have been retained and can be checked.

The author has reviewed the entire content sentence by sentence and takes full responsibility for the scientific validity and accuracy of this paper. The generative AI is not listed as an author and took no part in conceiving the research, formulating hypotheses, or judging conclusions.